\documentclass[conference]{IEEEtran}
\IEEEoverridecommandlockouts
\usepackage{cite}
\usepackage{amsmath,amssymb,amsfonts}
\usepackage{algorithmic}
\usepackage{graphicx}
\usepackage{textcomp}
\usepackage{xcolor}
\usepackage{subcaption}
\usepackage{booktabs}
\def\BibTeX{{\rm B\kern-.05em{\sc i\kern-.025em b}\kern-.08em
    T\kern-.1667em\lower.7ex\hbox{E}\kern-.125emX}}

\usepackage{xspace}
\newcommand*{\eg}{e.g.\@\xspace}
\newcommand*{\ie}{i.e.\@\xspace}
\newcommand*{\etal}{\emph{et al.}\@\xspace}

\usepackage[normalem]{ulem}

\begin{document}
\title{Recursive Training for \\Zero-Shot Semantic Segmentation}


\author{\IEEEauthorblockN{Ce Wang$^{1}$, Moshiur Farazi$^{2,1}$, Nick Barnes$^{1}$}
\IEEEauthorblockA{\textit{$^{1}$CECS, Australian National University}, Canberra, Australia  \\
\textit{$^{2}$Data61 -- CSIRO}, Canberra, Australia\\
\texttt{\{ce.wang1, moshiur.farazi, nick.barnes\}@anu.edu.au}}
}

\maketitle

\begin{abstract}
General purpose semantic segmentation relies on a backbone CNN network to extract discriminative features that help classify each image pixel into a `seen' object class (\ie, the object classes available during training) or a background class.
Zero-shot semantic segmentation is a challenging task that requires a computer vision model to identify image pixels belonging to an object class which it has never seen before.
Equipping a general purpose semantic segmentation model to separate image pixels of `unseen' classes from the background remains an open challenge.
Some recent models have approached this problem by fine-tuning the final pixel classification layer of a semantic segmentation model for a Zero-Shot setting, but struggle to learn discriminative features due to the lack of supervision.
We propose a recursive training scheme to supervise the retraining of a semantic segmentation model for a zero-shot setting using a pseudo-feature representation. 
To this end, we propose a Zero-Shot Maximum Mean Discrepancy (ZS-MMD) loss that weighs high confidence outputs of the pixel classification layer as a pseudo-feature representation, and feeds it back to the generator.
  By closing-the-loop on the generator end, we provide supervision during retraining that in turn helps the model learn a more discriminative feature representation for `unseen' classes. 
We show that using our recursive training and ZS-MMD loss, our proposed model achieves state-of-the-art performance on the Pascal-VOC 2012 dataset and Pascal-Context dataset.
\end{abstract}


\section{Introduction}
Semantic segmentation and zero-shot learning are both challenging computer vision tasks where the former requires a model to classify each image pixel and in the latter the model's training is restricted so that some test classes are not present during training. The task of zero shot semantic segmentation merges these two difficult tasks, and requires a model to classify each image pixel into \emph{seen} classes (\ie, test classes with training label) and \emph{unseen} classes (\ie, test classes without any training label). 

There have been significant advances in semantic segmentation and zero shot learning over the last few years. Recent semantic segmentation approaches leverage fully convolutional neural network architectures established by image classification models (\eg, AlexNet~\cite{krizhevsky2017imagenet}, ResNet~\cite{he2016deep}) as `encoders' to capture a coarse representation of the image; and then upsample the coarse feature map to the original pixel resolution via a `decoder', generating a semantic map.  A variety of approaches has been proposed to design better encoders and decoders, and recent works like DeepLabv3+~\cite{chen2018encoder}, U-net~\cite{ronneberger2015u} and Zoph \etal \cite{zoph2020rethinking} have achieved impressive performance on the semantic segmentation task. On the other hand, zero-shot learning has gained a lot of attention and most of its recent advances stem from learning a multimodal projection from image feature space to semantic space \cite{bucher2016improving, zhang2015zero}. The jointly embedded features in the multimodal space are then used for different zero-shot tasks \eg, classification~\cite{akata2015label, bucher2017generating}, detection~\cite{bansal2018zero, rahman2018deep}, recognition~\cite{morgado2017semantically, fu2018recent, wang2018zero}. These models suffer from bias towards seen classes, as the model focuses only on seen classes during training. \cite{cacheux2019modeling, rahman2018polarity} remove this bias from the perspective of loss function. \cite{bucher2017generating, kumar2018generalized, xian2018feature} solve this problem by using a generator to generate synthetic samples of unseen classes and use them to train the model. Although this removes the bias to some extent, their generators are still trained in the zero-shot setting, so these generators still have a bias towards seen classes, making the generated unseen samples inaccurate. 

\begin{figure}[t!]
\begin{center}
\includegraphics[width=\linewidth]{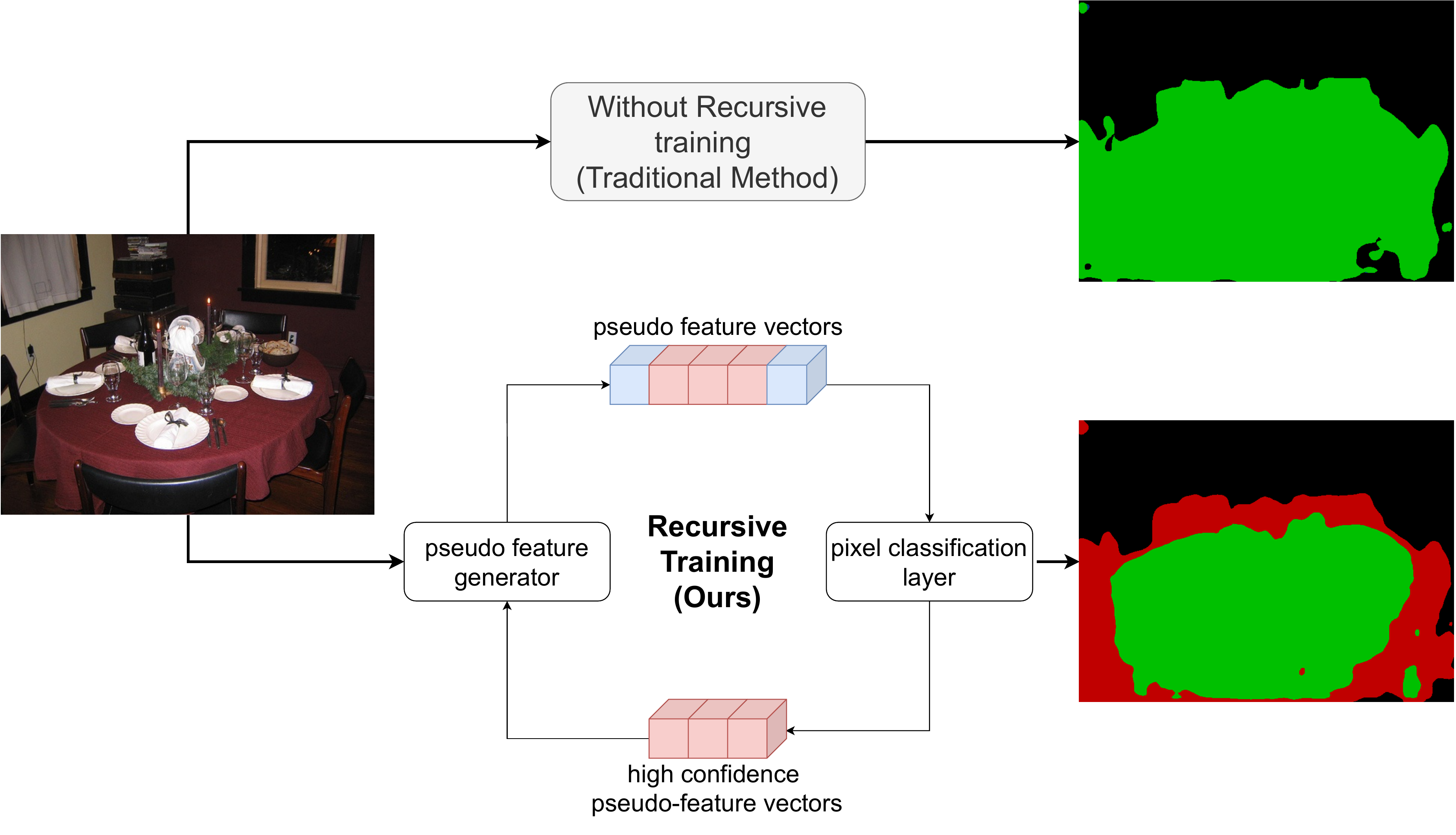}
\end{center}
\caption{\textbf{Recursive training improves Zero-Shot Semantic Segmentation performance.} By recursively training the feature generator with high confidence pseudo features, our model is able to learn more discriminative feature to segment unseen classes (\ie, unseen class \texttt{chair} is missed).
}
\label{fig:teaser}
\end{figure}
Most semantic segmentation methods work in a supervised setting which means the training set contains all classes in the test set~\cite{chen2017deeplab, hu2018learning}. The closed-set assumption that all test classes should be available during training limits the application of semantic segmentation. A desirable characteristic of a semantic segmentation model would be to identify rare classes which have few examples during training. Recently, several one-shot and few-shot semantic segmentation~\cite{xian2019semantic, dong2018few, BMVC2017_167} methods have been proposed to be used in such an unsupervised manner where the model is able to learn about a rare class from one or few-examples available the training set. Orthogonal to these is the motivation of zero-shot semantic segmentation, where the task is classifying image pixels of unseen object classes during test time. Until now, only two papers that we are aware of address zero-shot semantic segmentation: one is Zero-Shot Semantic Segmentation (ZS3Net)~\cite{bucher2019zero}, the other is by Kato \etal\cite{kato2019zero}. ZS3Net shows stronger results and is most comparable with our proposed model. Zero-shot semantic segmentation has two main challenges: \emph{first,} learning a mapping function to jointly embed the image data of an unseen object with the semantic class label of the unseen classes in a joint embedding space, and \emph{second,} the lack supervision while generating such joint embedding features. The first point speaks to the difficulty of this task and can be solved by providing more semantic information about the unseen class \cite{xian2019semantic}. In this work, we focus on the second point, which speaks about the need for a supervision signal during training to guide the generation of joint feature embeddings for zero-shot semantic segmentation.

Zero-Shot Semantic Segmentation (ZS3Net)~\cite{bucher2019zero} uses a generator to generate pseudo features of unseen classes, and uses both intermediate features extracted by the backbone network and pseudo features generated by this generator to train the segmentation model. We argue in this paper that ZS3Net is not able to fully utilize the weights learned by intermediate layers to govern itself to learn better joint feature embeddings for unseen classes. We propose to give the feature generator extra supervision by recursive training with intermediate features representation learned by the pixel classification layer. Intuitively, after training the pixel classification layer for several epochs, it is able to generate discriminative feature vectors for classifying unseen classes from seen classes. Drawing a parallel with the features generated by visual feature extractors (\eg, ResNet), we dub the intermediate features generated by the pixel classification layer as pseudo-feature vectors. We calculate the classification confidence from the pseudo-feature vectors and select only the ones that help the model to correctly predict the classification label or have high confidence. These high-confidence pseudo feature vectors are fed back to the pixel classification layer to predict another set of pseudo-feature vectors recursively. We hypothesise that these high confidence pseudo-feature vectors represent a better abstraction of the visual vectors for both seen and unseen classes, and the recursive training would allow the model to generate a more discriminative feature representation for the final classification. 

As we treat the high-confidence feature vectors like features extracted by a backbone CNN network, we weigh these features by their corresponding classification confidence with our proposed Zero-Shot Maximum Mean Discrepancy (ZS-MMD) loss. Our loss formulation takes inspiration from Generative Adversarial Nets (GAN) \cite{goodfellow2014generative} but does not use a discriminator to judge whether a feature is pseudo or not, rather we use the output of the pixel classification layer to pick high-confidence feature vectors. The main reason behind this is the lack of ground truth labels for the unseen classes. Unseen classes, by definition, are not coupled with labels, thus if the generator treats them as such, the features prediction would be seriously biased. Rather, our proposed ZS-MMD loss can be thought of a special case of co-training~\cite{blum1998combining} where the generator and pixel classification layers help each other, but are not parallel.

To evaluate our hypothesis, we compare our method with the state-of-the-art approach, ZS3Net~\cite{bucher2019zero}, on two popular semantic segmentation datasets, Pascal-VOC 2012~\cite{everingham2015pascal} and Pascal-Context~\cite{mottaghi2014role}, and show that our method gives better results. As the problem of lacking supervision of unseen classes is fundamental for zero-shot learning, we expect our method can inspire other works beyond zero shot semantic segmentation. Our key contributions are as follows:
\begin{itemize}
    \item We design a new training method that makes use of the pixel classification layer's ability to provide supervision for zero shot semantic segmentation using a pseudo feature generator.
    \item We propose Zero-Shot Maximum Mean Discrepancy (ZS-MMD) loss, that weighs pseudo-feature vectors based on their classification confidence and recursively trains the model to generate more discriminative feature vectors for the unseen classes.
    \item We show our method demonstrates improved results over the state-of-the-art approach.
\end{itemize}

\section{Related Works}


Semantic segmentation can be regarded as a pixel-wise classification task.
Deep Convolutional Neural Networks (CNNs) have been the dominant approach for semantic segmentation since  \cite{long2015fully}. In \cite{long2015fully}, the final convolutional layer of a CNN trained for classification is connected to a pixel-wise prediction layer. The whole network is trained using pixel-wise labeled images.
Subsequent papers have improved the results (\eg, \cite{chen2018encoder}, U-net\cite{ronneberger2015u}, Segnet\cite{badrinarayanan2017segnet}, PSPNet\cite{zhao2017pyramid}, and DeepLabv3+ \cite{chen2017deeplab}).  \cite{chen2017deeplab} achieves high quality results and is commonly used as a baseline in Weakly Supervised Semantic Segmentation \cite{wang2020self}, a related task. We adopt it as our backbone feature extraction network.



Zero-shot learning requires a model to predict both seen and unseen classes. However, traditionally a model cannot be extended to work on more classes after training. To give the model this ability, early methods such as \cite{lampert2013attribute} extract attributes of images and use them to perform prediction. However, due to limited attributes, the capacity of this type of method is limited. A more powerful method is to use semantic embeddings to model the relations between classes. Commonly used semantic embedding models are word2vec\cite{mikolov2013distributed}, GloVe\cite{pennington2014glove} and BERT\cite{devlin2018bert}. 
However, this type of methods has a strong bias towards seen classes, as the model focuses only on seen classes during training. Loss functions like Triplet loss\cite{cacheux2019modeling}, and Polarity loss\cite{rahman2018polarity} can help address bias. \cite{bucher2017generating, kumar2018generalized, xian2018feature} use another approach, using a generator to generate synthetic samples of unseen classes, and include them during training.

\begin{figure*}
\begin{center}
\includegraphics[width=.67\linewidth]{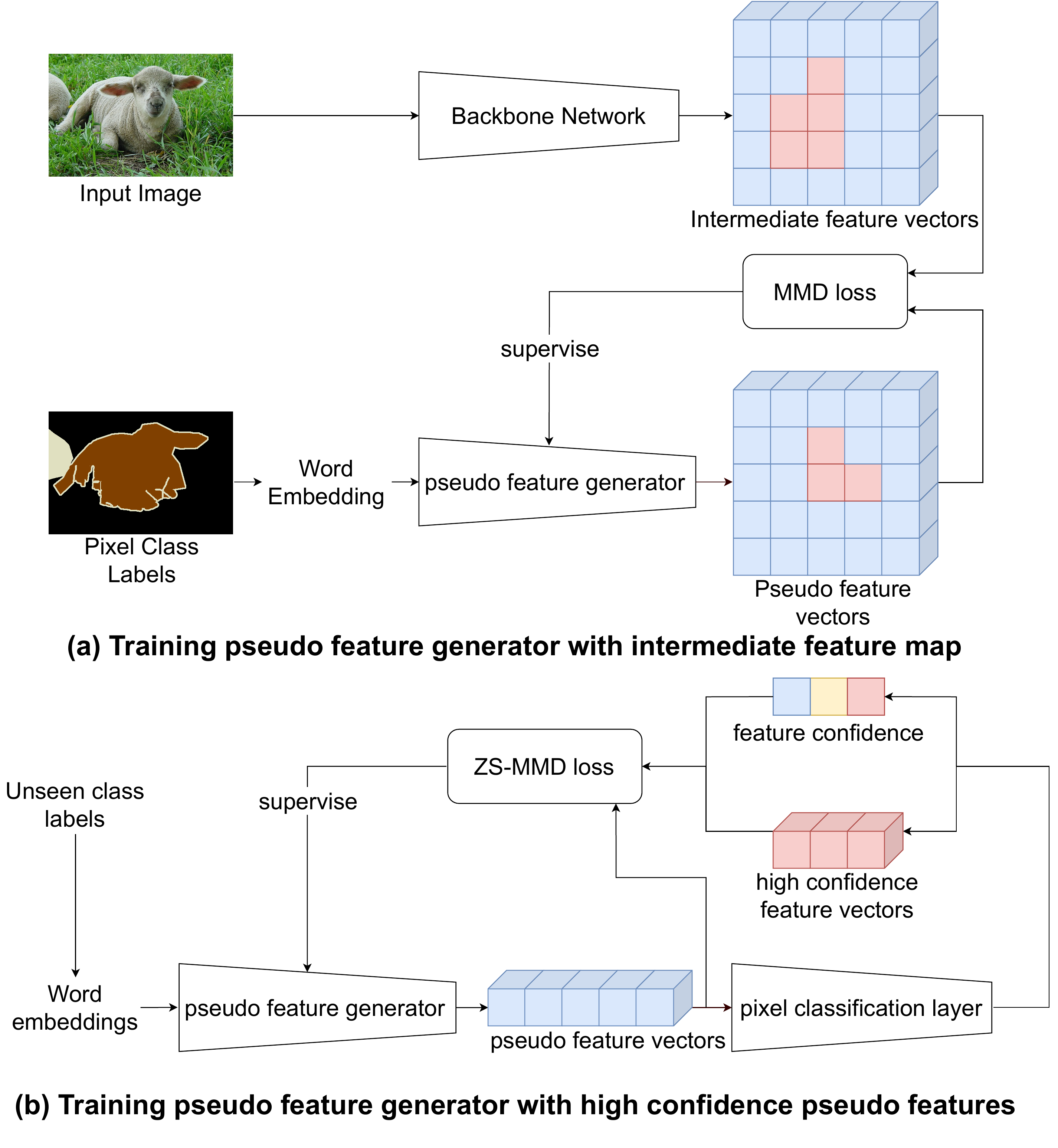}
\end{center}
\vspace{-1.5em}
\caption{\textbf{Overview of our proposed approach and training procedure.} (a) For seen classes, the pseudo-feature generator generates a pseudo-feature vector for each pixel. The ground truth is the intermediate feature vectors of all pixels extracted by the backbone network. For each seen class that appears in this image, we select the pseudo-feature vectors and intermediate feature vectors of pixels of this class, and compute the MMD loss, then use this to train the pseudo-feature generator. (b) When training with high-confidence pseudo-feature vectors, these are weighed by their confidence and serve as pseudo labels. The ZS-MMD loss guides the pseudo-feature generator to generate pseudo-feature vectors that are similar to those that make the classification result correct and of high confidence.}
\label{fig:architecture}
\end{figure*}

\section{Method}
\label{sec:method}

Our method is inspired by ZS3Net\cite{bucher2019zero}, where we introduce additional supervision to the pseudo-feature generator on how to generate pseudo-feature vectors from unseen classes. We first train our proposed model following the original training procedure proposed by ZS3Net\cite{bucher2019zero}. As discussed before, after a few epochs of training, the pixel classification layer has some ability to classify pixels from the unseen classes using the feature vectors corresponding to these pixels. We therefore make use of this ability 
to train the pseudo-feature generator. To make this method more effective, we modify the MMD Loss\cite{li2015generative} to take account of the confidence of high-confidence pseudo-feature vectors that serve as the pseudo-labels. We dub the modified MMD loss as Zero Shot MMD loss. 

\subsection{Network Architecture}
We use DeeplabV3+\cite{chen2018encoder} as our backbone network to extract pixel-wise feature vectors from images. Passing an image to the backbone network results in  an $H \times W \times 256$ dimension feature map, where $H$ and $W$ are the height and width of the input image, and each pixel has a feature vector of $256$ dimensions. This feature map is fed into the final $1 \times 1$ convolution pixel classification layer. This layer maps each 256-dimensional feature vector to a different number of classes depending on the dataset used. To enable the model to classify pixels from unseen classes, a Generative Moment Matching Network (GMMN)~\cite{li2015generative} is used as a pseudo-feature generator. This pseudo-feature generator uses 300-dimensional word2vec word embeddings and Gaussian noise of matching dimensionality to generate pseudo-feature vectors. Both of them are trained using the seen classes' word embeddings and intermediate feature vectors extracted by the backbone network. After training with seen classes, the pseudo-feature generator can generate pseudo-feature vectors given the word embeddings of the unseen classes. With such pseudo-feature vectors, we enable the pixel classification layer to classify pixels belonging to unseen classes. Our network architecture is illustrated in Fig.~\ref{fig:architecture}

\subsection{Recursive Training}
\label{sec:recursive training}

We pass the pseudo-feature vectors to the pixel classification layer to get the classification results and their corresponding confidence. Here, the confidence is obtained by applying the softmax function to the output of the pixel classification layer which is necessarily the estimated probability that a pixel belongs to a class. The pseudo-feature vectors corresponding to correct class predictions and with high probability are be used as pseudo-labels to train the pseudo feature generator. However, as these are pseudo-feature vectors, they should not be treated with equal weight in the loss to those computed when training with intermediate features. When using high-confidence pseudo-feature vectors to train the pseudo-feature generator, we empirically define a threshold, $\tau$, that decides high confidence and low confidence, and a factor $\gamma$ to reduce the loss computed during recursive training. We propose the ZS-MMD loss in the next section to facilitate the recursive training.

\subsection{ZS-MMD Loss}
The Generative Moment Matching Network (GMMN)~\cite{li2015generative} is  a neural network that generates samples having the same or a similar distribution as the training set. The loss function used is the MMD function, defined as follows:

\begin{equation}
\begin{aligned}
    L_{MMD^2}= & \frac{1}{N^{2}} \sum_{i=1}^{N} \sum_{i^{\prime}=1}^{N} k\left(x_{i}, x_{i^{\prime}}\right) \\
    & -\frac{2}{N M} \sum_{i=1}^{N} \sum_{j=1}^{M} k\left(x_{i}, y_{j}\right) \\
    & +\frac{1}{M^{2}} \sum_{j=1}^{M} \sum_{j^{\prime}=1}^{M} k\left(y_{j}, y_{j^{\prime}}\right)
    \end{aligned}
\end{equation}
\noindent
where $x_i$, $x_{i^{\prime}}$, $y_i$ and $y_{i^{\prime}}$ are data from the training set and the generated samples, 
$M$ is the number of samples in the training set, 
$N$ is the number of generated samples
and $k$ is the kernel function. When using high-confidence pseudo-feature vectors as pseudo-labels to train the pseudo-feature generator, it is intuitive that pseudo-feature vectors with higher confidence should weigh more, so, we propose the ZS-MMD loss that can weigh each sample differently. For the context of training the pseudo-feature generator, we use the classification confidence mentioned in \ref{sec:recursive training} to weigh high-confidence pseudo-feature vectors. The ZS-MMD loss function is defined as follows:

\begin{equation}
\begin{aligned}
    L_{ZS-MMD^2}= & \frac{1}{{\left(\sum_{i} c_{i}\right)}^{2}} \sum_{i=1}^{Q} \sum_{i^{\prime}=1}^{Q} c_{i} c_{i^{\prime}} k\left(a_{i}, a_{i^{\prime}}\right) \\
    & -\frac{2}{P \sum_{i} c_{i}} \sum_{i=1}^{Q} \sum_{j=1}^{P} c_{i} k\left(a_{i}, b_{j}\right) \\
    & +\frac{1}{P^{2}} \sum_{j=1}^{P} \sum_{j^{\prime}=1}^{P} k\left(b_{j}, b_{j^{\prime}}\right)
    \end{aligned}
\end{equation}
where $c_i$ denotes the confidence corresponding to the i-th high-confidence pseudo-feature vector (i.e., $c_i > \tau$), $b$ denotes the generated samples, and $a$ the high confidence generated samples, $P$ denotes the number of generated samples, and $Q$ denotes the number of generated samples where $c_i > \tau$.

\begin{table*}[ht!]
\begin{center}
\caption{\textbf{Segmentation Performance on Pascal-VOC 2012 dataset.} We report \textbf{K} = 2, 4, 6, 8, 10 unseen classes. We select the first 2, 4, 6, 8, 10 classes of \texttt{cow, motorbike, airplane, sofa, cat, tv, train, bottle, chair, potted-plant} as unseen classes in Pascal-VOC 2012 dataset, and report pixel accuracy (PA), mean accuracy (MA), mean intersection-over-union (mIoU) of seen and unseen classes, and corresponding overall performance with harmonic mean of mIoU (hIoU).}
\label{table:voc results}
\begin{tabular}{llllllllllllll}
\toprule
 & & \multicolumn{3}{c}{Seen} & & \multicolumn{3}{c}{Unseen} & & \multicolumn{4}{c}{Overall} \\
\cline{3-5} \cline{7-9} \cline{11-14}
\textbf{K} & Model & PA & MA & mIoU & & PA & MA & mIoU & & PA & MA & mIoU & hIoU \\
\hline\hline
 & ZS3Net & 93.6 & 84.9 & 72.0 & & 52.8 & 53.7 & 35.4 & & 92.7 & 81.9 & 68.5 & 47.5 \\
2 & Ours & \textbf{94.0} & 84.2 & 71.6 & & \textbf{54.4} & \textbf{54.2} & \textbf{37.5} & & \textbf{92.8} & 80.1 & 67.7 & \textbf{49.2} \\
\hline
 & ZS3Net & 92.0 & 78.3 & 66.4 & & 43.1 & 45.7 & 23.2 & & 89.8 & 72.1 & 58.2 & 34.4 \\
4 & Ours & \textbf{93.6} & \textbf{80.1} & \textbf{68.9} & & 40.4 & 43.6 & \textbf{27.0} & & \textbf{91.1} & \textbf{73.3} & \textbf{60.7} & \textbf{38.8} \\
\hline
 & ZS3Net & 85.5 & 52.1 & 47.3 & & 67.3 & 60.7 & 24.2 & & 84.2 & 54.6 & 40.7 & 32.0 \\
6 & Ours & \textbf{90.7} & \textbf{59.9} & \textbf{51.1} & & 63.9 & 57.3 & \textbf{25.5} & & \textbf{87.7} & \textbf{56.5} & \textbf{43.6} & \textbf{33.6} \\
\hline
 & ZS3Net & 81.6 & 31.6 & 29.2 & & 68.7 & 62.3 & 22.9 & & 80.3 & 43.3 & 26.8 & 25.7 \\
8 & Ours & \textbf{85.8} & \textbf{37.6} & \textbf{32.3} & & 62.3 & 55.7 & \textbf{25.2} & & \textbf{83.4} & \textbf{44.5} & \textbf{29.6} & \textbf{28.3} \\
\hline
 & ZS3Net & 82.7 & 37.4 & 33.9 & & 55.2 & 45.7 & 18.1 & & 79.6 & 41.4 & 26.3 & 23.6 \\
10 & Ours & 82.0 & \textbf{38.3} & \textbf{34.4} & & \textbf{60.0} & \textbf{55.9} & \textbf{23.9} & & 79.5 & \textbf{46.7} & \textbf{29.4} & \textbf{28.2} \\
\bottomrule
\end{tabular}
\end{center}
\end{table*}
\begin{table*}
\begin{center}
\caption{\textbf{Segmentation Performance on Pascal-Context dataset.} We select the first \textbf{K} = 2, 4, 6, 8, 10 classes of \texttt{cow, motorbike, airplane, sofa, cat, tv, train, bottle, chair, potted-plant} as unseen classes in Pascal-Context dataset, and report pixel accuracy (PA), mean accuracy (MA), mean intersection-over-union (mIoU) of seen and unseen classes, and corresponding overall performance with harmonic mean of mIoU (hIoU).}
\label{table:context results}
\begin{tabular}{llllllllllllll}
\toprule
 & & \multicolumn{3}{c}{Seen} & & \multicolumn{3}{c}{Unseen} & & \multicolumn{4}{c}{Overall} \\
\cline{3-5} \cline{7-9} \cline{11-14}
\textbf{K} & Model & PA & MA & mIoU & & PA & MA & mIoU & & PA & MA & mIoU & hIoU \\
\hline\hline
 & ZS3Net & 71.6 & 52.4 & 41.6 & & 49.3 & 46.2 & 21.6 & & 71.2 & 52.2 & 41.0 & 28.4 \\
2 & Ours & \textbf{71.7} & 51.9 & 41.3 & & 43.7 & 41.3 & 21.2 & & 71.2 & 51.6 & 40.6 & 28.0 \\
\hline
 & ZS3Net & 68.4 & 46.1 & 37.2 & & 58.4 & 53.3 & 24.9 & & 67.8 & 46.6 & 36.4 & 29.8 \\
4 & Ours & \textbf{68.7} & \textbf{49.8} & \textbf{38.7} & & 52.4 & 50.5 & \textbf{25.9} & & 67.6 & \textbf{49.8} & \textbf{37.9} & \textbf{31.0} \\
\hline
 & ZS3Net & 63.3 & 38.0 & 32.1 & & 63.6 & 55.8 & 20.7 & & 63.3 & 39.8 & 30.9 & 25.2 \\
6 & Ours & \textbf{68.4} & \textbf{44.8} & \textbf{35.9} & & 48.9 & 43.0 & \textbf{21.5} & & \textbf{66.7} & \textbf{44.6} & \textbf{34.4} & \textbf{26.9} \\
\hline
 & ZS3Net & 51.4 & 23.9 & 20.9 & & 68.2 & 59.9 & 16.0 & & 53.1 & 28.7 & 20.3 & 18.1 \\
8 & Ours & \textbf{56.9} & \textbf{28.9} & \textbf{24.2} & & 57.5 & 47.6 & 15.9 & & \textbf{56.9} & \textbf{31.4} & \textbf{23.0} & \textbf{19.2} \\
\hline
 & ZS3Net & 53.5 & 23.8 & 20.8 & & 58.6 & 43.2 & 12.7 & & 52.8 & 27.0 & 19.4 & 15.8 \\
10 & Ours & \textbf{55.2} & \textbf{26.4} & \textbf{22.6} & & 52.5 & 39.6 & \textbf{12.8} & & \textbf{54.9} & \textbf{28.6} & \textbf{21.0} & \textbf{16.3} \\

\bottomrule
\end{tabular}
\end{center}
\end{table*}

\section{Experiments}

\subsection{Experimental Details}
\label{sec:exp details}
The baseline to our proposed method is ZS3Net\cite{bucher2019zero}. We use backbone DeeplabV3+~\cite{chen2018encoder} based on ResNet101\cite{he2016deep} for extracting pixel-wise feature vectors, pre-trained on the ImageNet~\cite{imagenet_cvpr09} and fine-tuned on the seen classes. After fine-tuning, the weights of the backbone network is frozen. Only the final pixel classification layer and the pseudo-feature generator are trained. They are trained simultaneously, which means, given an image, we use the backbone network to extract pixel-wise feature vectors, then, we use the feature vectors of the seen classes as pseudo-labels to train the pseudo-feature generator. If this image contains unseen classes, for the pixels corresponding to these classes would be used to train the pseudo-feature generator. Subsequently, based on whether this image contains unseen classes or not, generated pseudo-feature vectors or intermediate feature vectors extracted by the backbone network are used to train the final pixel classification layer. 

During training, the images are resized so that their short sides have 312 pixels, and their aspect ratio is preserved. During validation, they are resized so that their short sides have 513 pixels. Data augmentation including random flipping, random cropping, and Gaussian blur are also utilized when training our model. The segmentation model is trained using the SGD \cite{bottou2010large} optimizer, with a learning rate of $1e^{-7}$, weight decay of $5e^{-4}$ and momentum of 0.9. The structure of the pseudo-feature generator is the same as described in ZS3Net paper \cite{bucher2019zero}, trained using Adam optimizer \cite{kingma2014adam} with a learning rate of $2e^{-4}$. The batch-size for the segmentation model was set to 8 images, and the batch-size for the pseudo feature generator was set to 128 feature vectors. The model is trained using a NVIDIA RTX2080ti GPU.

We evaluate our method on two datasets, Pascal-VOC 2012\cite{everingham2015pascal} and Pascal-Context\cite{mottaghi2014role}. For segmentation, Pascal-VOC 2012 provides 1464 training images and 1449 validation images, and annotations for 20 classes. Pascal-Context provides full segmentation annotations for Pascal-VOC 2010\cite{everingham2015pascal}. It provides 4,998 training images and 5105 validation images, and annotations for 59 classes. To compare with our baseline, we also make use of semantic boundary annotations\cite{hariharan2011semantic} to help training. The experiments were done with 2, 4, 6, 8, 10 unseen classes. For both experiment, the unseen classes are the first 2, 4, 6, 8, 10 of [cow, motorbike, airplane, sofa, cat, tv, train, bottle, chair, potted-plant]. For Pascal-VOC 2012\cite{everingham2015pascal}, when training the pseudo-feature generator, $\tau$ and $\gamma$ for 4, 6, 8 unseen classes are empirically chosen to be $0.7$ and $\frac{1}{80}$, and for 2 and 10 unseen classes, $0.85$ and $\frac{1}{40}$. For Pascal-Context\cite{mottaghi2014role}, $\tau$ and $\gamma$
for 2, 4, 6, 8, 10 unseen classes are $[0.85, \frac{1}{160}]$, $[0.7, \frac{1}{160}]$, $[0.85, \frac{1}{80}]$, $[0.85, \frac{1}{80}]$, $[0.7, \frac{1}{160}]$. For both datasets, we do not experiment with self-training, which has been explored in ZS3Net \cite{bucher2019zero}, as self-training allows the model to receive information about the unseen classes and this breaks the setting of zero-shot learning \cite{xian2017zero}. Same as ZS3Net \cite{bucher2019zero}, we use pixel accuracy (PA), mean accuracy (MA), mean intersection-over-union (mIoU) and harmonic mean of mIoU (hIoU) of seen and unseen classes as evaluation metrics.

\begin{figure*}
\begin{center}
    \begin{subfigure}[b]{0.23\textwidth}
        \centering
        \caption{Input Image}
        \includegraphics[width=\textwidth]{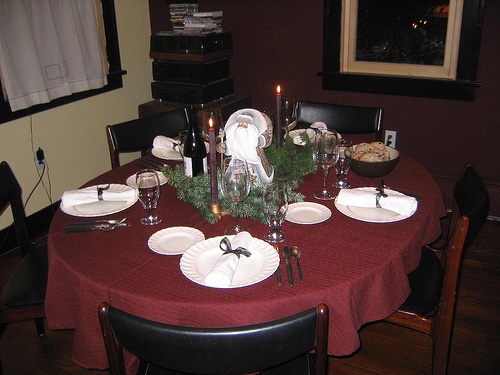}
        \rule{0pt}{0ex}  
        \includegraphics[width=\textwidth]{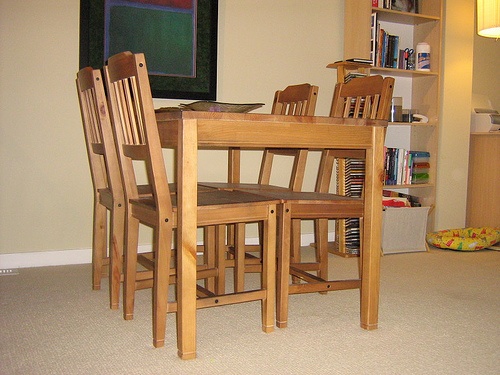}
        \rule{0pt}{0ex}
        \includegraphics[width=\textwidth]{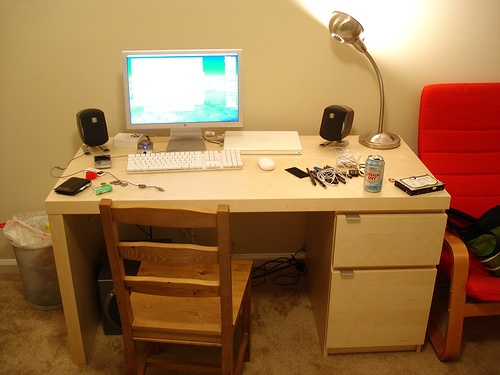}
        \rule{0pt}{0ex}
        \includegraphics[width=\textwidth]{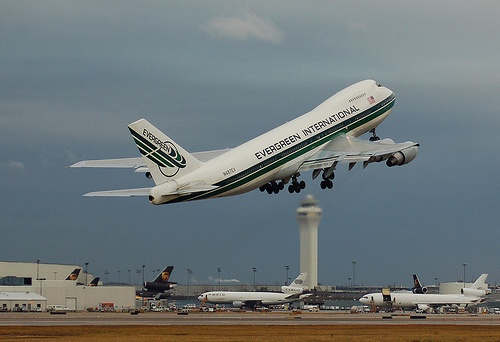}
        \rule{0pt}{0ex}
        \includegraphics[width=\textwidth]{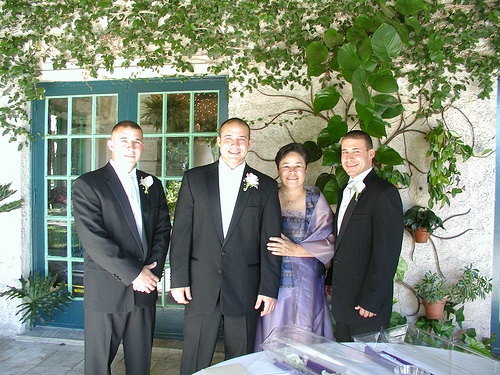}
        \rule{0pt}{0ex}
        \includegraphics[width=\textwidth]{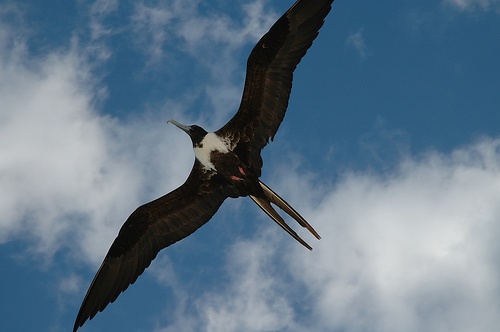}

        \label{table_img}
    \end{subfigure}
    \begin{subfigure}[b]{0.23\textwidth}
        \centering
        \caption{Ground Truth}
        \includegraphics[width=\textwidth]{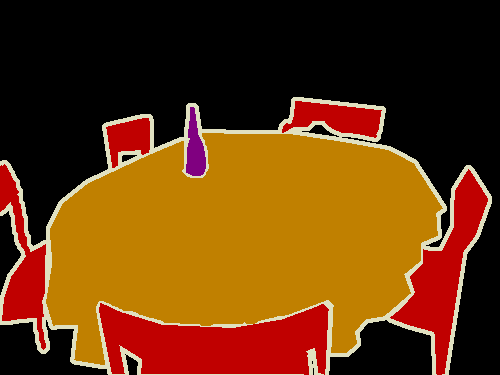}
        \rule{0pt}{0ex}  
        \includegraphics[width=\textwidth]{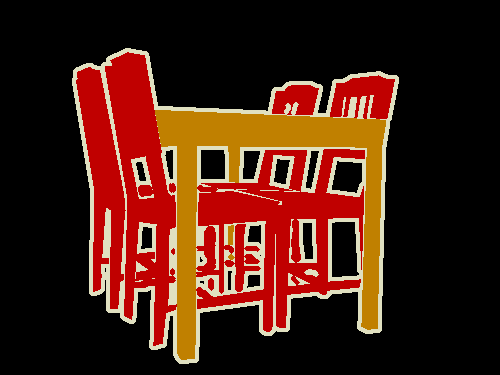}
        \rule{0pt}{0ex} 
        \includegraphics[width=\textwidth]{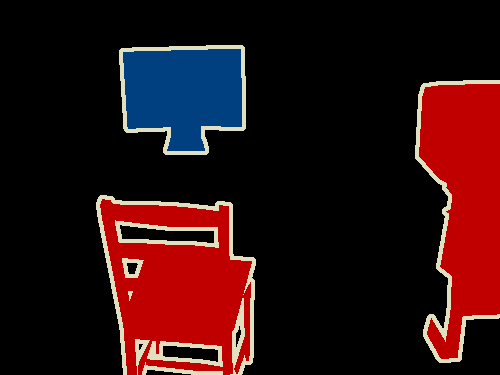}
        \rule{0pt}{0ex}
        \includegraphics[width=\textwidth]{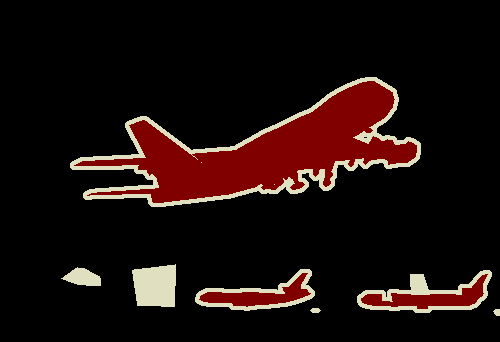}
        \rule{0pt}{0ex}  
        \includegraphics[width=\textwidth]{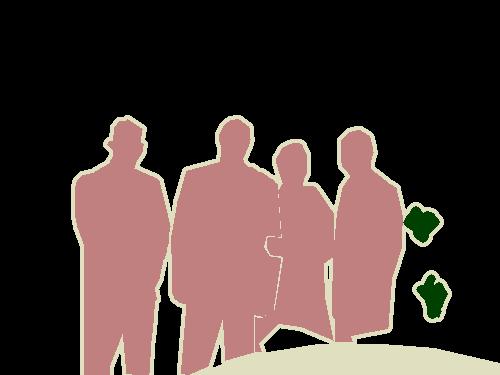}
        \rule{0pt}{0ex}
        \includegraphics[width=\textwidth]{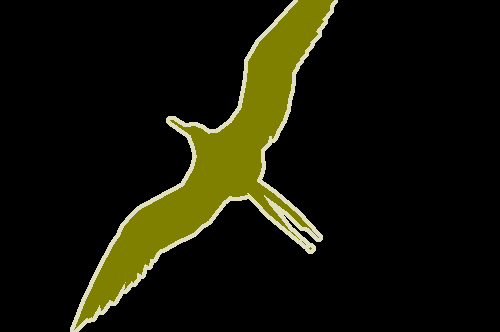}

        \label{table_gt}
    \end{subfigure}
    \begin{subfigure}[b]{0.23\textwidth}
        \centering
        \caption{ZS3Net~\cite{bucher2019zero}}
        \includegraphics[width=\textwidth]{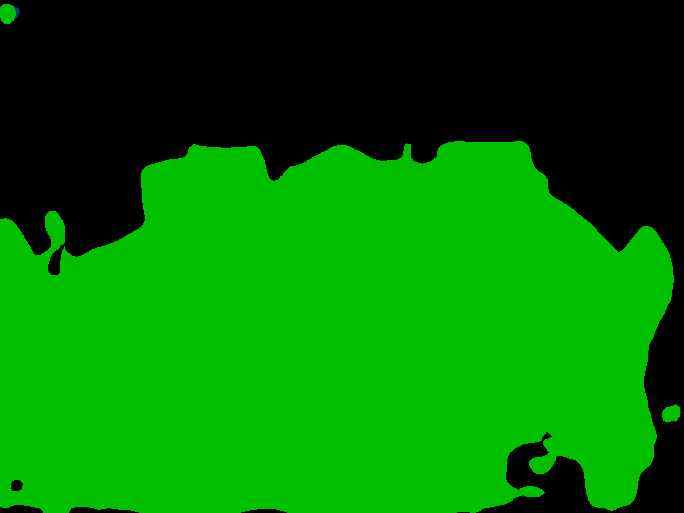}
        \rule{0pt}{0ex}  
        \includegraphics[width=\textwidth]{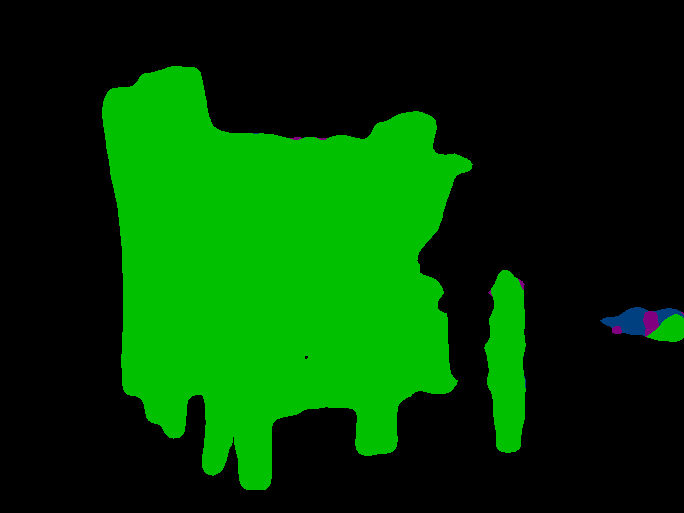}
        \rule{0pt}{0ex}
        \includegraphics[width=\textwidth]{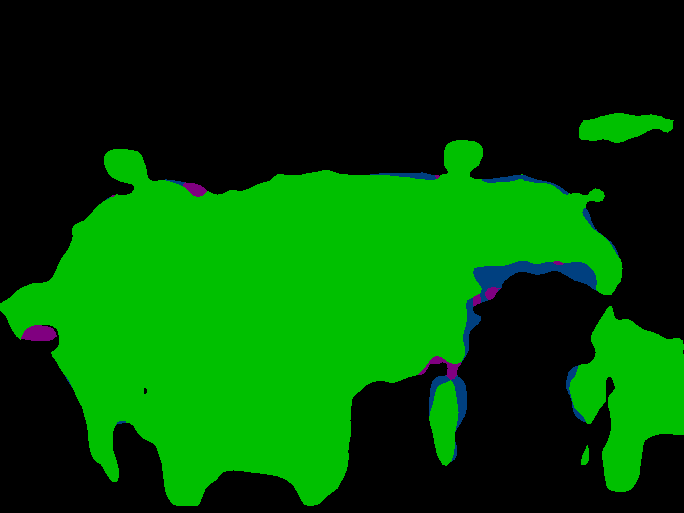}
        \rule{0pt}{0ex}
        \includegraphics[width=\textwidth]{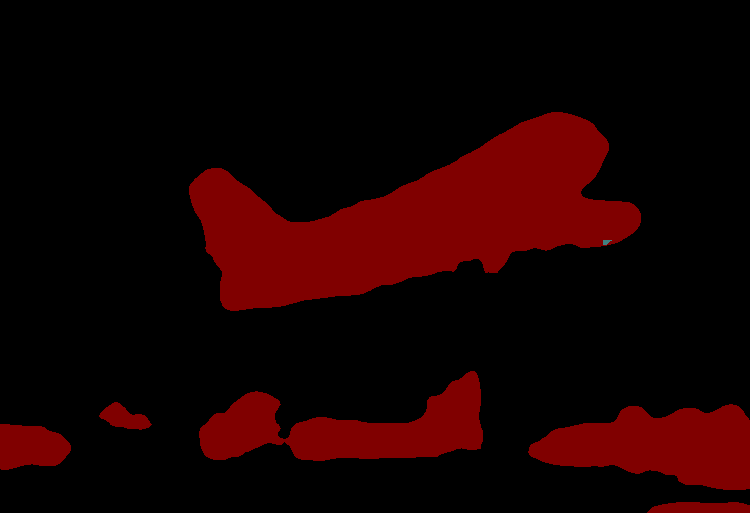}
        \rule{0pt}{0ex}  
        \includegraphics[width=\textwidth]{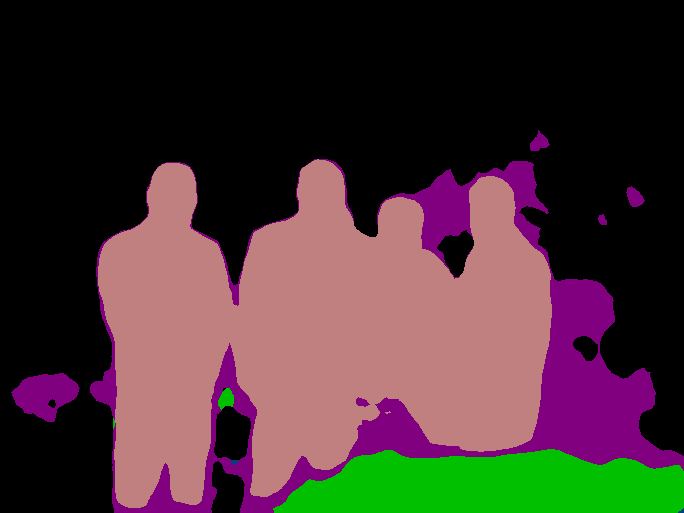}
        \rule{0pt}{0ex}
        \includegraphics[width=\textwidth]{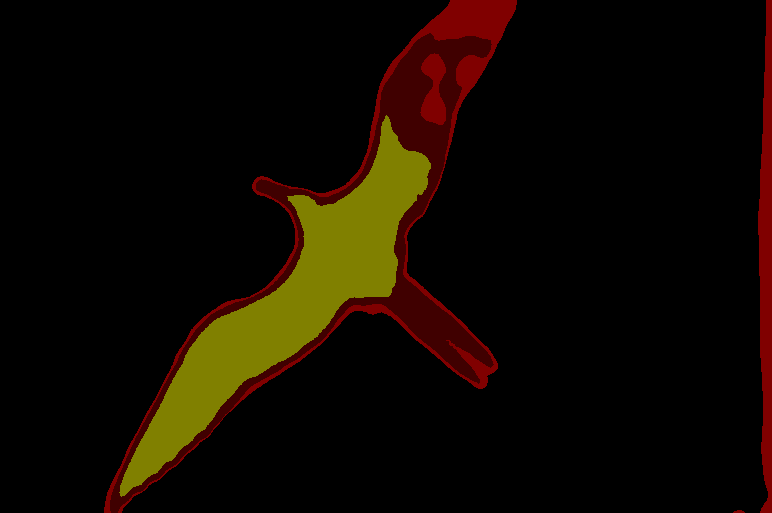}

        \label{table_bs}
    \end{subfigure}
    \begin{subfigure}[b]{0.23\textwidth}
        \centering
        \caption{Ours}        
        \includegraphics[width=\textwidth]{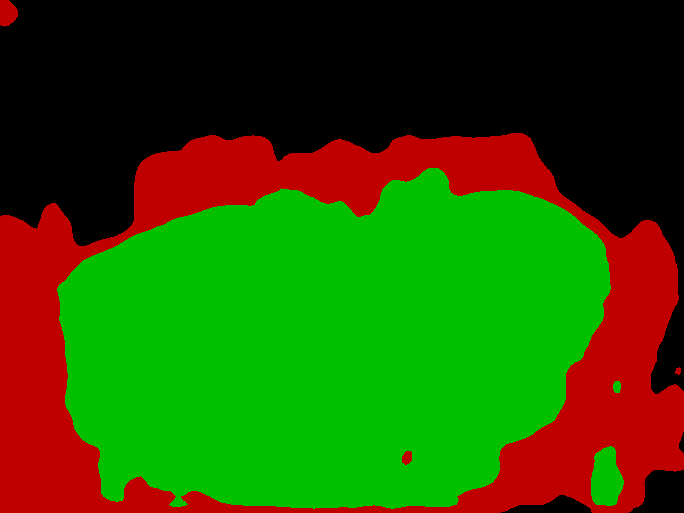}
        \rule{0pt}{0ex}  
        \includegraphics[width=\textwidth]{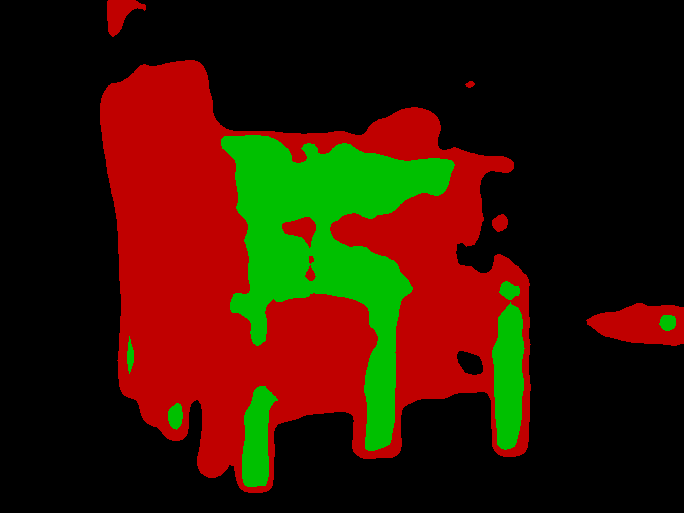}
        \rule{0pt}{0ex}
        \includegraphics[width=\textwidth]{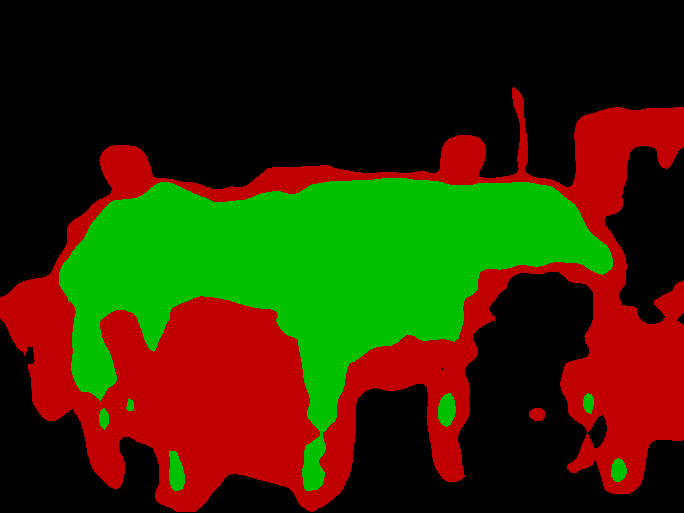}
        \rule{0pt}{0ex}
        \includegraphics[width=\textwidth]{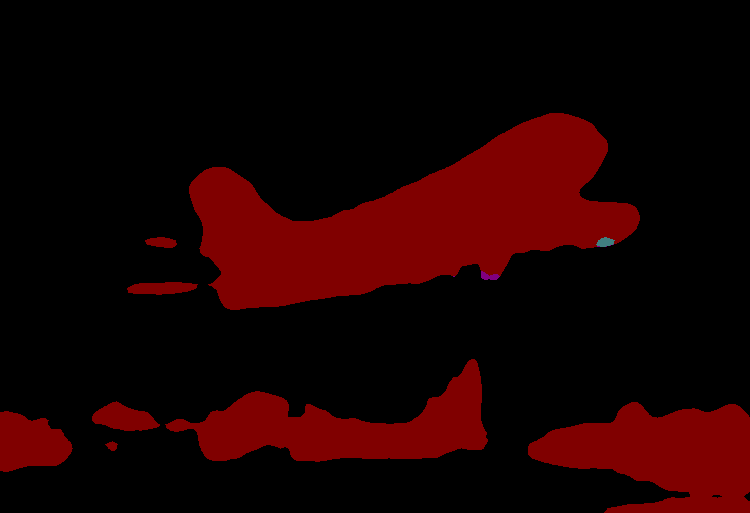}
        \rule{0pt}{0ex}  
        \includegraphics[width=\textwidth]{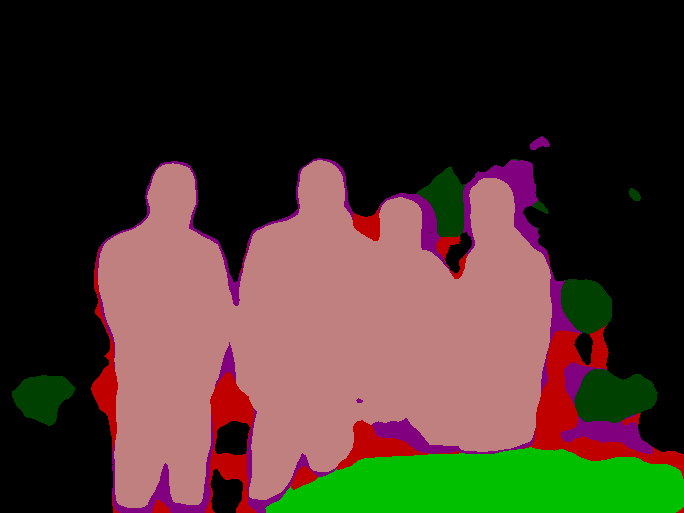}
        \rule{0pt}{0ex}
        \includegraphics[width=\textwidth]{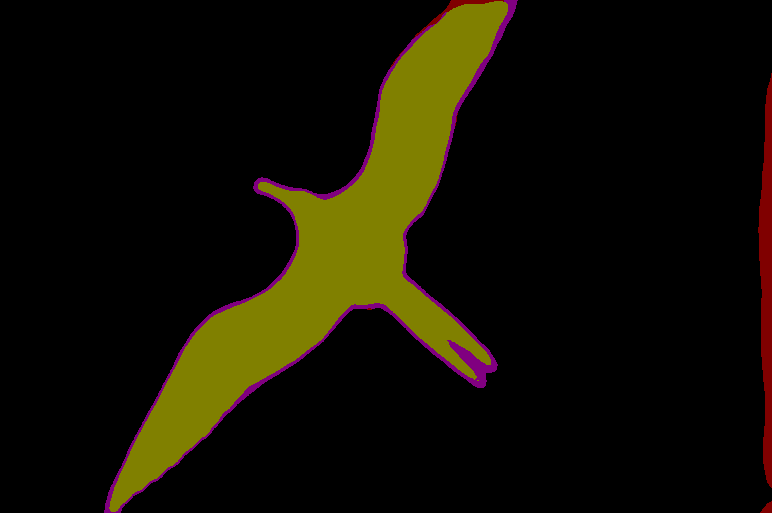}

        \label{table_my}
    \end{subfigure}
\end{center}
\caption{\textbf{Qualitative results on Pascal-VOC 2012 dataset comparing our proposed model with the state-of-the-art.} Both models are trained on 10 unseen classes described in Sec.~\ref{sec:exp details}. It can be seen that the models can better segment both unseen classes (\eg, \texttt{chair, airplane}) and seen classes (\eg, \texttt{person, bird}).}
\label{fig:voc samples}
\end{figure*}

\begin{table*}
\begin{center}
\caption{\textbf{Ablation Study.} Comparison between weighing the pseudo-feature vectors equally and weighing the pseudo-feature vectors using their corresponding confidence.}
\label{table:ablation}
\begin{tabular}{llllllllllllll}
\toprule
 & & \multicolumn{3}{c}{Seen} & & \multicolumn{3}{c}{Unseen} & & \multicolumn{4}{c}{Overall} \\
\cline{3-5} \cline{7-9} \cline{11-14}
\textbf{K} & Model & PA & MA & mIoU & & PA & MA & mIoU & & PA & MA & mIoU & hIoU \\
\hline\hline
 & Equal Weight & 93.8 & 82.8 & 71.1 & & 47.3 & 47.7 & 32.0 & & 92.7 & 80.2 & 67.8 & 44.1 \\
2 & Final & 94.0 & 84.2 & 71.6 & & 54.4 & 54.2 & 37.5 & & 92.8 & 80.1 & 67.7 & 49.2 \\
\hline
 & Equal Weight & 93.6 & 80.1 & 66.9 & & 31.2 & 35.0 & 22.0 & & 90.8 & 72.2 & 58.3 & 33.1 \\
4 & Final & 93.6 & 80.1 & 68.9 & & 40.4 & 43.6 & 27.0 & & 91.1 & 73.3 & 60.7 & 38.8 \\
\hline
 & Equal Weight & 85.9 & 51.3 & 47.3 & & 67.8 & 60.7 & 23.8 & & 84.6 & 54.0 & 40.1 & 31.7 \\
6 & Final & 90.7 & 59.9 & 51.1 & & 63.9 & 57.3 & 25.5 & & 87.7 & 56.5 & 43.6 & 33.6 \\
\hline
 & Equal Weight & 87.3 & 47.8 & 41.2 & & 58.9 & 51.1 & 24.9 & & 84.5 & 49.1 & 35.0 & 31.0 \\
8 & Final & 85.8 & 37.6 & 32.3 & & 62.3 & 55.7 & 25.2 & & 83.4 & 44.5 & 29.6 & 28.3 \\
\hline
 & Equal Weight & 89.1 & 56.2 & 45.5 & & 48.8 & 44.1 & 21.2 & & 84.5 & 50.4 & 34.0 & 28.9 \\
10 & Final & 82.0 & 38.3 & 34.4 & & 60.0 & 55.9 & 23.9 & & 79.5 & 46.7 & 29.4 & 28.2 \\
\bottomrule
\end{tabular}
\end{center}
\end{table*}

\subsection{Pascal-VOC 2012}
Tab.~\ref{table:voc results} shows the performance of our model and our baseline ZS3Net\cite{bucher2019zero}, on \textbf{K} unseen classes. We see that our model gives better intersection-over-union (IoU) than ZS3Net on most cases, for both seen and unseen classes. Especially, for \textbf{K} = 2 and 10, the PA, MA and mIoU scores are significantly better than the baseline for unseen classes. Further, for all cases of \textbf{K}, our model achieves better mIoU than the baseline for unseen classes and in overall (seen and unseen combined). A similar trend can be observed for seen classes except for \textbf{K} = 2, where our model is slightly outperformed by the baseline in MA and mIoU metric. We can also see some cases where the PA and MA of our model is lower than ZS3Net, we argue that IoU is a more robust measure as PA and MA can be impacted by modifying the recall rate.

\subsection{Pascal-Context}
Tab.~\ref{table:context results} shows the performance of our model and ZS3Net\cite{bucher2019zero} on \textbf{K} unseen classes. The ZS3Net model used graph-context encoding to provide context-conditioned pseudo-feature vectors and improved their results. However, when using this method, one cluster of pixels that has the same label are considered as a whole, therefore, the number of embeddings for an image shrinks to the number of clusters in this image. This number is too small to be used in our method. For example, if there is only one high confidence pseudo-feature vector, using it as ground truth will tell the pseudo feature generator that this pseudo-feature vector is definitely correct, then the gradient will explode. Therefore, in Tab.~\ref{table:context results}, we compare our results with the results of ZS3Net obtained without graph-context encoding. We can see that for \textbf{K} = 4, 8, 10, the mIoU on unseen classes of our method are better than that of ZS3Net. For \textbf{K} = 2 and 6, although our method's mIoU on unseen classes is lower than that of ZS3Net, the difference is small. For performance on seen classes, aside from the results on \textbf{K} = 2, our model provides better results for all metrics, which further demonstrates that our method is more robust. This robustness further results in better overall performance, where we show that for \textbf{K}=4, 6, 8, 10, the overall performance of our method is clearly better than ZS3Net.

\subsection{Qualitative Results}
We compare qualitative results of our proposed model with the state-of-the-art ZS3Net\cite{bucher2019zero} model in Fig. \ref{fig:voc samples}. The segmentation masks reported are generated by our model and ZS3Net, trained on 10 unseen classes (\textbf{K} = 10). We show six segmentation results from Pascal-VOC 2012\cite{everingham2015pascal} combining both unseen and seen classes. In Fig.~\ref{fig:voc samples}, rows 1-3, the input image contains instances of \texttt{table, chair} and the ground truth segmentation mask. \texttt{chair} is an unseen class and we can see that the ZS3Net cannot differentiate between the generated segmentation mask, hence the green mask combining instances of both \texttt{table} and \texttt{chair} (row 1-3, col 3). However, our proposed method can reasonably distinguish between them and separate them in green and red segmentation masks (row 1-3, col 4). For another unseen class, \texttt{airplane}, in row 4, the result generated by our model contains more detail than ZS3Net, as it can segment the tail of the largest airplane better. For rows 5-6, we show that our model is also able to estimate better segmentation masks for seen classes. In rows 5, we see that both ZS3Net and our proposed model picks up seen class \texttt{person}, however our model is able to segment unseen objects from the background better than ZS3Net (\eg, missing \texttt{potted-plant} in row 5, col 4). Furthermore, in row 6, our model is able to predict the whole bird with its wings as a single segmentation mask compared its counterpart (row 6, col 3 vs. 4). Its worth mentioning that, when we set \textbf{K} = 10 for Pascal-VOC 2012 dataset, its a hard test setting as out of 21 segmentation classes, almost half are set to be unseen. This has a negative influence on the image variety resulting in poor performance from ZS3Net, particularly for unseen objects. Even in this setting, when nearly half of the object classes are unavailable during training, our method can still generate segmentation masks that are clear and reasonably closer to the ground truth.

\subsection{Ablation Study}
We perform an ablation study to evaluate if selecting high confidence pseudo-features compared to setting equal weights for all pseudo features can give better semantic segmentation performance. We show results for this experiment for all \textbf{K} on the Pascal-VOC 2012 dataset in Tab.~\ref{table:ablation}. While training our model, we first we select all pseudo-features and weigh them equally (\ie, `Equal Weight' rows in Tab.~\ref{table:ablation}) and compare their performance against weighted high confidence pseudo features version of our model (\ie, `Final' rows in Tab.~\ref{table:ablation}). We observe that for most cases, weighing the selected feature vectors with confidence gives better mIoU scores on unseen classes. Therefore, with this ablation we provide support for our hypothesis that weighing pseudo feature vectors based on confidence score leads to better performance, which is a key component for our method.


\section{Conclusion:}
In this work, we proposed a recursive training procedure for the Zero Shot semantic segmentation. With our proposed ZS-MMD loss, we make use of the pixel classification layers ability to generate discriminative feature representation, by iterative generating high-confidence pseudo feature vectors, which in turn allows the model to better segment unseen and seen classes. With extensive quantitative and qualitative experimentation on two popular semantic segmentation dataset, each with five different settings, we showcase the effectiveness of our propose approach. Further, we perform ablation to show that weighing the pseudo-features vectors based on their confidence, give better semantic segmentation performance. Although, in the scope of this paper, we only experimented with zero-shot semantic segmentation, we believe that our recursive training formulation is generic, and can be extended to other zero-shot tasks.


\bibliographystyle{IEEEtran}
\bibliography{egbib}

\begin{thebibliography}{10}
\providecommand{\url}[1]{#1}
\csname url@samestyle\endcsname
\providecommand{\newblock}{\relax}
\providecommand{\bibinfo}[2]{#2}
\providecommand{\BIBentrySTDinterwordspacing}{\spaceskip=0pt\relax}
\providecommand{\BIBentryALTinterwordstretchfactor}{4}
\providecommand{\BIBentryALTinterwordspacing}{\spaceskip=\fontdimen2\font plus
\BIBentryALTinterwordstretchfactor\fontdimen3\font minus
  \fontdimen4\font\relax}
\providecommand{\BIBforeignlanguage}[2]{{%
\expandafter\ifx\csname l@#1\endcsname\relax
\typeout{** WARNING: IEEEtran.bst: No hyphenation pattern has been}%
\typeout{** loaded for the language `#1'. Using the pattern for}%
\typeout{** the default language instead.}%
\else
\language=\csname l@#1\endcsname
\fi
#2}}
\providecommand{\BIBdecl}{\relax}
\BIBdecl

\bibitem{krizhevsky2017imagenet}
A.~Krizhevsky, I.~Sutskever, and G.~E. Hinton, ``Imagenet classification with
  deep convolutional neural networks,'' \emph{Communications of the ACM},
  vol.~60, no.~6, pp. 84--90, 2017.

\bibitem{he2016deep}
K.~He, X.~Zhang, S.~Ren, and J.~Sun, ``Deep residual learning for image
  recognition,'' in \emph{Proceedings of the IEEE conference on computer vision
  and pattern recognition}, 2016, pp. 770--778.

\bibitem{chen2018encoder}
L.-C. Chen, Y.~Zhu, G.~Papandreou, F.~Schroff, and H.~Adam, ``Encoder-decoder
  with atrous separable convolution for semantic image segmentation,'' in
  \emph{Proceedings of the European conference on computer vision (ECCV)},
  2018, pp. 801--818.

\bibitem{ronneberger2015u}
O.~Ronneberger, P.~Fischer, and T.~Brox, ``U-net: Convolutional networks for
  biomedical image segmentation,'' in \emph{International Conference on Medical
  image computing and computer-assisted intervention}.\hskip 1em plus 0.5em
  minus 0.4em\relax Springer, 2015, pp. 234--241.

\bibitem{zoph2020rethinking}
B.~Zoph, G.~Ghiasi, T.-Y. Lin, Y.~Cui, H.~Liu, E.~D. Cubuk, and Q.~V. Le,
  ``Rethinking pre-training and self-training,'' \emph{arXiv preprint
  arXiv:2006.06882}, 2020.

\bibitem{bucher2016improving}
M.~Bucher, S.~Herbin, and F.~Jurie, ``Improving semantic embedding consistency
  by metric learning for zero-shot classiffication,'' in \emph{European
  Conference on Computer Vision}.\hskip 1em plus 0.5em minus 0.4em\relax
  Springer, 2016, pp. 730--746.

\bibitem{zhang2015zero}
Z.~Zhang and V.~Saligrama, ``Zero-shot learning via semantic similarity
  embedding,'' in \emph{Proceedings of the IEEE international conference on
  computer vision}, 2015, pp. 4166--4174.

\bibitem{akata2015label}
Z.~Akata, F.~Perronnin, Z.~Harchaoui, and C.~Schmid, ``Label-embedding for
  image classification,'' \emph{IEEE transactions on pattern analysis and
  machine intelligence}, vol.~38, no.~7, pp. 1425--1438, 2015.

\bibitem{bucher2017generating}
M.~Bucher, S.~Herbin, and F.~Jurie, ``Generating visual representations for
  zero-shot classification,'' in \emph{Proceedings of the IEEE International
  Conference on Computer Vision Workshops}, 2017, pp. 2666--2673.

\bibitem{bansal2018zero}
A.~Bansal, K.~Sikka, G.~Sharma, R.~Chellappa, and A.~Divakaran, ``Zero-shot
  object detection,'' in \emph{Proceedings of the European Conference on
  Computer Vision (ECCV)}, 2018, pp. 384--400.

\bibitem{rahman2018deep}
S.~Rahman and S.~Khan, ``Deep multiple instance learning for zero-shot image
  tagging,'' in \emph{Asian Conference on Computer Vision}.\hskip 1em plus
  0.5em minus 0.4em\relax Springer, 2018, pp. 530--546.

\bibitem{morgado2017semantically}
P.~Morgado and N.~Vasconcelos, ``Semantically consistent regularization for
  zero-shot recognition,'' in \emph{Proceedings of the IEEE Conference on
  Computer Vision and Pattern Recognition}, 2017, pp. 6060--6069.

\bibitem{fu2018recent}
Y.~Fu, T.~Xiang, Y.-G. Jiang, X.~Xue, L.~Sigal, and S.~Gong, ``Recent advances
  in zero-shot recognition: Toward data-efficient understanding of visual
  content,'' \emph{IEEE Signal Processing Magazine}, vol.~35, no.~1, pp.
  112--125, 2018.

\bibitem{wang2018zero}
X.~Wang, Y.~Ye, and A.~Gupta, ``Zero-shot recognition via semantic embeddings
  and knowledge graphs,'' in \emph{Proceedings of the IEEE conference on
  computer vision and pattern recognition}, 2018, pp. 6857--6866.

\bibitem{cacheux2019modeling}
Y.~L. Cacheux, H.~L. Borgne, and M.~Crucianu, ``Modeling inter and intra-class
  relations in the triplet loss for zero-shot learning,'' in \emph{Proceedings
  of the IEEE International Conference on Computer Vision}, 2019, pp.
  10\,333--10\,342.

\bibitem{rahman2018polarity}
S.~Rahman, S.~Khan, and N.~Barnes, ``Polarity loss for zero-shot object
  detection,'' \emph{arXiv preprint arXiv:1811.08982}, 2018.

\bibitem{kumar2018generalized}
V.~Kumar~Verma, G.~Arora, A.~Mishra, and P.~Rai, ``Generalized zero-shot
  learning via synthesized examples,'' in \emph{Proceedings of the IEEE
  conference on computer vision and pattern recognition}, 2018, pp. 4281--4289.

\bibitem{xian2018feature}
Y.~Xian, T.~Lorenz, B.~Schiele, and Z.~Akata, ``Feature generating networks for
  zero-shot learning,'' in \emph{Proceedings of the IEEE conference on computer
  vision and pattern recognition}, 2018, pp. 5542--5551.

\bibitem{chen2017deeplab}
L.-C. Chen, G.~Papandreou, I.~Kokkinos, K.~Murphy, and A.~L. Yuille, ``Deeplab:
  Semantic image segmentation with deep convolutional nets, atrous convolution,
  and fully connected crfs,'' \emph{IEEE transactions on pattern analysis and
  machine intelligence}, vol.~40, no.~4, pp. 834--848, 2017.

\bibitem{hu2018learning}
R.~Hu, P.~Doll{\'a}r, K.~He, T.~Darrell, and R.~Girshick, ``Learning to segment
  every thing,'' in \emph{Proceedings of the IEEE Conference on Computer Vision
  and Pattern Recognition}, 2018, pp. 4233--4241.

\bibitem{xian2019semantic}
Y.~Xian, S.~Choudhury, Y.~He, B.~Schiele, and Z.~Akata, ``Semantic projection
  network for zero-and few-label semantic segmentation,'' in \emph{Proceedings
  of the IEEE/CVF Conference on Computer Vision and Pattern Recognition}, 2019,
  pp. 8256--8265.

\bibitem{dong2018few}
N.~Dong and E.~P. Xing, ``Few-shot semantic segmentation with prototype
  learning.'' in \emph{BMVC}, vol.~3, no.~4, 2018.

\bibitem{BMVC2017_167}
\BIBentryALTinterwordspacing
A.~Shaban, S.~Bansal, Z.~Liu, I.~Essa, and B.~Boots, ``One-shot learning for
  semantic segmentation,'' in \emph{Proceedings of the British Machine Vision
  Conference (BMVC)}, G.~B. Tae-Kyun~Kim, Stefanos~Zafeiriou and
  K.~Mikolajczyk, Eds.\hskip 1em plus 0.5em minus 0.4em\relax BMVA Press,
  September 2017, pp. 167.1--167.13. [Online]. Available:
  \url{https://dx.doi.org/10.5244/C.31.167}
\BIBentrySTDinterwordspacing

\bibitem{bucher2019zero}
M.~Bucher, V.~Tuan-Hung, M.~Cord, and P.~P{\'e}rez, ``Zero-shot semantic
  segmentation,'' in \emph{Advances in Neural Information Processing Systems},
  2019, pp. 468--479.

\bibitem{kato2019zero}
N.~Kato, T.~Yamasaki, and K.~Aizawa, ``Zero-shot semantic segmentation via
  variational mapping,'' in \emph{Proceedings of the IEEE International
  Conference on Computer Vision Workshops}, 2019, pp. 0--0.

\bibitem{goodfellow2014generative}
I.~Goodfellow, J.~Pouget-Abadie, M.~Mirza, B.~Xu, D.~Warde-Farley, S.~Ozair,
  A.~Courville, and Y.~Bengio, ``Generative adversarial nets,'' in
  \emph{Advances in neural information processing systems}, 2014, pp.
  2672--2680.

\bibitem{blum1998combining}
A.~Blum and T.~Mitchell, ``Combining labeled and unlabeled data with
  co-training,'' in \emph{Proceedings of the eleventh annual conference on
  Computational learning theory}, 1998, pp. 92--100.

\bibitem{everingham2015pascal}
M.~Everingham, S.~A. Eslami, L.~Van~Gool, C.~K. Williams, J.~Winn, and
  A.~Zisserman, ``The pascal visual object classes challenge: A
  retrospective,'' \emph{International journal of computer vision}, vol. 111,
  no.~1, pp. 98--136, 2015.

\bibitem{mottaghi2014role}
R.~Mottaghi, X.~Chen, X.~Liu, N.-G. Cho, S.-W. Lee, S.~Fidler, R.~Urtasun, and
  A.~Yuille, ``The role of context for object detection and semantic
  segmentation in the wild,'' in \emph{Proceedings of the IEEE Conference on
  Computer Vision and Pattern Recognition}, 2014, pp. 891--898.

\bibitem{long2015fully}
J.~Long, E.~Shelhamer, and T.~Darrell, ``Fully convolutional networks for
  semantic segmentation,'' in \emph{Proceedings of the IEEE conference on
  computer vision and pattern recognition}, 2015, pp. 3431--3440.

\bibitem{badrinarayanan2017segnet}
V.~Badrinarayanan, A.~Kendall, and R.~Cipolla, ``Segnet: A deep convolutional
  encoder-decoder architecture for image segmentation,'' \emph{IEEE
  transactions on pattern analysis and machine intelligence}, vol.~39, no.~12,
  pp. 2481--2495, 2017.

\bibitem{zhao2017pyramid}
H.~Zhao, J.~Shi, X.~Qi, X.~Wang, and J.~Jia, ``Pyramid scene parsing network,''
  in \emph{Proceedings of the IEEE conference on computer vision and pattern
  recognition}, 2017, pp. 2881--2890.

\bibitem{wang2020self}
Y.~Wang, J.~Zhang, M.~Kan, S.~Shan, and X.~Chen, ``Self-supervised equivariant
  attention mechanism for weakly supervised semantic segmentation,'' in
  \emph{Proceedings of the IEEE/CVF Conference on Computer Vision and Pattern
  Recognition}, 2020, pp. 12\,275--12\,284.

\bibitem{lampert2013attribute}
C.~H. Lampert, H.~Nickisch, and S.~Harmeling, ``Attribute-based classification
  for zero-shot visual object categorization,'' \emph{IEEE transactions on
  pattern analysis and machine intelligence}, vol.~36, no.~3, pp. 453--465,
  2013.

\bibitem{mikolov2013distributed}
T.~Mikolov, I.~Sutskever, K.~Chen, G.~S. Corrado, and J.~Dean, ``Distributed
  representations of words and phrases and their compositionality,'' in
  \emph{Advances in neural information processing systems}, 2013, pp.
  3111--3119.

\bibitem{pennington2014glove}
J.~Pennington, R.~Socher, and C.~D. Manning, ``Glove: Global vectors for word
  representation,'' in \emph{Proceedings of the 2014 conference on empirical
  methods in natural language processing (EMNLP)}, 2014, pp. 1532--1543.

\bibitem{devlin2018bert}
J.~Devlin, M.-W. Chang, K.~Lee, and K.~Toutanova, ``Bert: Pre-training of deep
  bidirectional transformers for language understanding,'' \emph{arXiv preprint
  arXiv:1810.04805}, 2018.

\bibitem{li2015generative}
Y.~Li, K.~Swersky, and R.~Zemel, ``Generative moment matching networks,'' in
  \emph{International Conference on Machine Learning}, 2015, pp. 1718--1727.

\bibitem{imagenet_cvpr09}
J.~Deng, W.~Dong, R.~Socher, L.-J. Li, K.~Li, and L.~Fei-Fei, ``Imagenet: A
  large-scale hierarchical image database,'' in \emph{2009 IEEE conference on
  computer vision and pattern recognition}.\hskip 1em plus 0.5em minus
  0.4em\relax Ieee, 2009, pp. 248--255.

\bibitem{bottou2010large}
L.~Bottou, ``Large-scale machine learning with stochastic gradient descent,''
  in \emph{Proceedings of COMPSTAT'2010}.\hskip 1em plus 0.5em minus
  0.4em\relax Springer, 2010, pp. 177--186.

\bibitem{kingma2014adam}
D.~P. Kingma and J.~Ba, ``Adam: A method for stochastic optimization,''
  \emph{arXiv preprint arXiv:1412.6980}, 2014.

\bibitem{hariharan2011semantic}
B.~Hariharan, P.~Arbel{\'a}ez, L.~Bourdev, S.~Maji, and J.~Malik, ``Semantic
  contours from inverse detectors,'' in \emph{2011 International Conference on
  Computer Vision}.\hskip 1em plus 0.5em minus 0.4em\relax IEEE, 2011, pp.
  991--998.

\bibitem{xian2017zero}
Y.~Xian, B.~Schiele, and Z.~Akata, ``Zero-shot learning-the good, the bad and
  the ugly,'' in \emph{Proceedings of the IEEE Conference on Computer Vision
  and Pattern Recognition}, 2017, pp. 4582--4591.

\end{thebibliography}

\end{document}